# PBM: A NEW DATASET FOR BLOG MINING

**MEHWISH AZIZ**
FAST – National University of
Computer & Emerging Sciences

**MUHAMMAD RAFI**
FAST – National University of
Computer & Emerging Sciences

*ABSTRACT*
Text mining is becoming vital as Web 2.0 offers collaborative content creation and sharing. Now Researchers have growing interest in text mining methods for discovering knowledge. Text mining researchers come from variety of areas like: Natural Language Processing, Computational Linguistic, Machine Learning, and Statistics. A typical text mining application involves preprocessing of text, stemming and lemmatization, tagging and annotation, deriving knowledge patterns, evaluating and interpreting the results. There are numerous approaches for performing text mining tasks, like: clustering, categorization, sentimental analysis, and summarization. There is a growing need to standardize the evaluation of these tasks. One major component of establishing standardization is to provide standard datasets for these tasks. Although there are various standard datasets available for traditional text mining tasks, but there are very few and expensive datasets for blog-mining task. Blogs, a new genre in web 2.0 is a digital diary of web user, which has chronological entries and contains a lot of useful knowledge, thus offers a lot of challenges and opportunities for text mining. In this paper, we report a new indigenous dataset for Pakistani Political Blogosphere. The paper describes the process of data collection, organization, and standardization. We have used this dataset for carrying out various text mining tasks for blogosphere, like: blog-search, political sentiments analysis and tracking, identification of influential blogger, and clustering of the blog-posts. We wish to offer this dataset free for others who aspire to pursue further in this domain.

## 1 INTRODUCTION

The emergence of Web 2.0 raised standard associated with web applications. These standards are proposed to support variant services for information sharing, interoperability, collaboration, and user-centered applications. Web 2.0 web sites are capable of allowing users to interact and contribute in web sites' content as user-generated content. Web 2.0 sites offers the most adaptable internet knowledge creation processes, which triggers many internet virtual community based hosted services, social networking sites, wikis, and blogs etc. The main subject of this paper is Blogs. Blogs-online digital diary like application helps users to connect in the virtual community by creating, reviewing, commenting or tagging the user-generated content/ blog post. Blog post is simply a web page with updates in its contents and archives of previously created content linked with other blog posts of other users under the same category. This sharing of blogs by different users over the same categories fall under one big umbrella termed as Blogosphere—at times helping 'awareness building social entities' to bring important issues in the lime light. Blogosphere's blogs differs from other collections of documents in internet on three levels: content, structure, and timeline. The Blogosphere is indeed a collection of all blogs, which is further sub-divided to different genres like sports, politics, media, and culture. The most popular among all these genres are media and politics. Media blogs are related to fashion, music, actors and movies. Political blogs are uncensored political bias over a political issue posted by an individual who may be an ordinary human or a news person or may be a political leader himself/herself. In comparison to media related blogs, influence of political blogs have strong implications when online community reviews them. Opinion by a journalist, analyst, politician or a common man may shift the thinking pattern of users three sixty degrees. Political blogs is one the most popular categories in blogosphere. It has generated a power of opinion that is experienced several times in different parts of the world. One of the most famous real-life examples is the US Presidential Election of 2004 and 2008. Prior to these elections, bloggers of political blogs created the opinion in blog posts. These opinions not only affected the online community but also the mainstream media. The mainstream media utilized the bloggers' opinion to cover top dominant politicians in the Presidential Election. One famous case-study of political blogosphere is [1], which reports this phenomenon. It also reports that there was a distinct difference in the blogging behavior of liberal and conservatives. The importance of political blogs, blogosphere and bloggers is realized since then. The dataset reported in this paper, comprises of a subset of political blogosphere of Pakistani politics. The geopolitical importance of Pakistan is now widely accepted and its role in war against terrorism is greatly acknowledged. This political blog-dataset has been utilized in our previous research on topics like: Blog Search [3], Blog-Similarity [4], finding influence bloggers [5], opinion mining and tracking. A dataset is essential to carry out research in the realm of blogosphere. The major contribution of this paper is collecting, processing, standardizing and publishing a dataset of political blogosphere of Pakistani Politics. The rest of the paper is organized as follow: in section 2, we discussed the similar work from other researchers; in section 3, dataset construction process; in section 4, discussion of the application we developed so far based on this dataset. In the last section, we report the conclusion and possible future work.

## 2. RELATED WORK

The collections of data for the sake of performing data/text mining experiments have a long history. In the area of text mining specifically for blog mining task the first attempt to form a collection of blog's dataset was made by [2], this collection was used for TREC 2006 blog-track which was introduced very first time in TREC conference in the same year. The cost of the TREC Blog06 dataset was about £200 to £500 which was way beyond the purchasing power of the University in the Third World. The cost of the data collection was the major reason that very few Universities has actually participated in this Track of TREC 2006. The dataset reported in this paper, which in deed was a motivation to prepare a dataset which can be offered free to researchers. Blogs-weblog is an online user's diary like application, where users keep their daily writing pertinent to any topic of his like/dislike. A special genre of blogs that comments on politics are termed as political blogs and a collection of all such political blogs comprises political blogosphere. The political blog-post usually has a clear-cut political bias. We have collected a subset of Pakistani Political blogosphere for the sake of building a dataset, where researchers can carry out research in the realm of political blog mining.  The research in political-blogosphere started way back in 2004, during the US Presidential Election [1], the paper discusses the collection of blogs from liberal and conservative communities, their interaction and overlap on various topics. It also collected 1000 blog-post of a single day, to capture and study the blog-rolls (cross link of bloggers). The study was carried out on a very limited dataset but still very valuable. It has been proved in [6] that blogs are very proactive in changing thoughts, making readers on agreement or on disagreement on certain issues, disseminating and echoing ones voice related to any matter. This study also used 1000 blog-posts from 33 of the world's top blogs. Opinion mining from blogs is another area where few of the researchers have developed few datasets, example of such dataset is [7]. A dataset that has been very recently used for detecting gender of bloggers is reported in [8]. The very latest and freely available dataset is from [9] International AAAI Conference on Weblog and Social Media (ICWSM), which has been used on a wide Variety of mining tasks. The dataset is provided by Spinn3r.com is a set of 44 million blog-posts made collected from August-October 2008. Our dataset is very different from this dataset, as our dataset contains only political blogs related to political issues of Pakistan. The ICWSM dataset covers a lot of big events occurred during the same duration like Olympics and US Presidential Nomination. Our dataset is more rigorous and homogeneous as far as topic of the research is concerned. The collected dataset is very rich on NEWS reported event and issues erupted from the news as well as from personal reviews on these NEWS. We believe that this dataset offers a lot of opportunity for researchers from social science and computer science domains.

## 3 DATASET CONSTRUCTION

In this section, we describe the experience of generating the dataset and the prime features of the dataset that can be used by the researchers. Also, the discussion will lead further to constraints that are existent in the dataset with reference to the identified features.

In order to help the researchers' in understanding what are the possible ways this dataset can be used for research, we will explain the key areas that we find it useful for. And, which areas of research have we worked on by using this dataset.

### 3.1 EXPERIENCE OF DATA GATHERING

The collection of RSS Feeds by using RSS Feeder started from March $8^{th}$, 2008 and continued till Sept $2^{nd}$, 2010 which brings about 14,115 blog posts and comments from referred blog sites. The collected blog posts are 9936 and comments collected over few blog sites are 4179. Covering all the blog posts in perspective of Pakistani Politics, we could not combine those blog sites which are originating for political discussion but with some other country's perspectives. Therefore, the collected blog sites seem to be limited.

The dataset is gathered by a pre-developed RSS Feeder [1] which allows to gather XML-Format RSS Feeds from any site. We selected only Pakistani Political blog sites. The top 5 most active sources used for blog posts collection is given in Tab.1 below.

Tab.1 Top 5 blog sites used for collection of blog posts with reference to Pakistani political issues only

| Serial No. | Blog Site | Number of Blog Posts |
|---|---|---|
| 1 | http://newsrss.bbc.co.uk/rss/newsonline_world_edition/south_asia/rss.xml | 2290 |
| 2 | http://feeds.feedburner.com/dawn/news/pakistan?format=xml | 1829 |
| 3 | http://feeds2.feedburner.com/blogspot/tzao | 1351 |
| 4 | http://feeds.feedburner.com/AwazApniBaatApni | 1035 |
| 5 | http://feeds2.feedburner.com/pakspectator/rhqq?format=xml | 964 |

### 3.2. FEATURES

The dataset so far gathered is stored in an MS-Access Database with the database design that has two relational tables titled as Source Channels and RSS Feeds. These tables are automatically created as soon as RSS Feeder is installed. Amongst the two tables, Channels are storing the Channel ID, Title, and FeedURL for defining the profile of the Feed Channels used as a source of data gathering. The source channel status is covered by Frequency, Last Updated, and Next Update is updated in the database whenever the aggregator pushes a request and source response is pulled out. The FolderPath is maintaining the source collected posts final path. The number of posts from the source is referred from ItemCount and UnreadCount. The second table is titled as RSSFeeds, and its fields like Guid and ChannelID refers to the source feed links.XML holds the entire blog post which includes Title, Publication Date, Source Links as well. But this information is separately provided as Title and PubDate by the aggregator as well. Also, the XML field is stored as <item> only but it holds the entire blog posts in RSSFeeds Table. This XML-Format data shows that title and data of the blog post is in English using Roman words as well. The blog post (XML data) also show that all outlinks and inlinks are also specified within actual blog content. The blog posts collected are of variant length with variant number of outlinks and inlinks used.

### 3.3 DATASET CONSTRAINTS

The dataset so far collected has fewer numbers of comments as all the feed links used do not have a separate comment source channels dedicated. The collected blog posts are not completely separated as individual fields like author name, total count of comments for each blog post. Also, the blog posts linked over the same title from the same source channel are not linked with each other.

### 3.4. DATASET ANALYSIS

This dataset is analyzed in terms of collection pattern from different blogs and the way authors publish blogs. Over the years from 2006 till 2010, posts collection from selected blog sites is shown in Fig.1 given below:

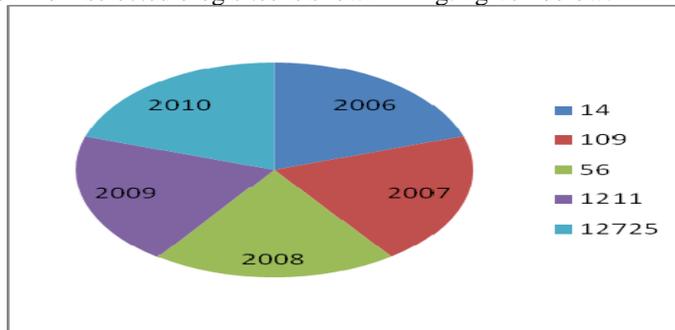

Fig.1 Year-wise posts coverage by bloggers over blog sites

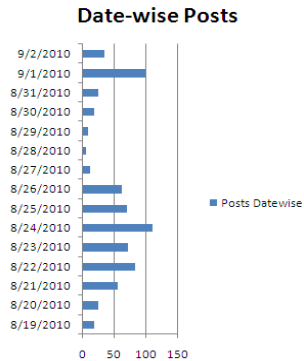 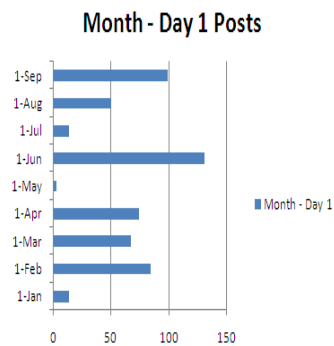

Fig.2 Blog Posts coverage on daily basis. Showing that bloggers don't follow any specific days to blog. In fact on an average it has been found that bloggers mostly post blogs on mid-weeks that is by Wednesday.

Fig.3 Each month's first day has been noticed to be used by bloggers to post more blogs either to pre-view the previous month's effects and their expected consequences in Pakistani politics.

### 3.5 USAGE

In blogs, each blog post hold individual's perception over particular issues. These perceptions are generally entailing positive, negative or neutral opinion. This influences the reviewers that can be identified in comments added over the blog post. These opinions can be used very effectively to identify the sentiments of an online community. Bloggers and their reviewers normally post an opinion in a blog or a comment in either direct or comparative manner. From direct manner, opinion can be semantically classified as positive, negative or neutral. And in comparative opinion, semantics can be classified as either good or bad. This extraction and analysis of opinion can be termed as Sentiment Analysis and Opinion Mining. Thus, opinion mining can be performed using this dataset specific to Pakistani Politics.

This dataset can also be used in blog search engines and blog ranking systems for Pakistani Political Blogs, as opinion retrieval and analysis plays a vital role to give results of a searched entity or judging relevancy of a blog post over many other blog posts on a particular topic.

## 4. OUR APPLICATIONS

We have used this dataset to test few criteria of semantic blog search. For this purpose, the dataset has been utilized to cluster the blogs into related categories based on semantic similarity measurement criteria. For clustering, initially we used an algorithm termed as BPSM [3] which utilized blog posts' title and their content to measure the sentence-wise semantic similarity amongst blog posts of various blog sites. We have also utilized the same dataset on identifying influential bloggers in the Pakistani political blogosphere using the semantic influence measuring criteria like posts' quality in terms of its length and number of outlinks along with the uniqueness of the posts' content. The algorithm we built to get the results by using this dataset is termed as SIIB [4]. Another criterion in which this dataset has been used is to ranks and search for semantically relevant blogs of influential bloggers by using the influence flow i.e. inlinks and outlinks along with the semantics of the blog posts.

To perform all the above tasks we have parsed XML blog posts content over part of this dataset collection (around 3778 from Nov 16$^{th}$, 2009 till Feb 12$^{th}$, 2010). This parsed data is stored further in the database as a table titled as Parsed RSSFeeds. This table provides a refinement of the RSS Feeds XML data into certain data fields which were used for performing text mining based techniques for different problems.

Few problems with the XML content identified involve data fields like; Bloggers, Comments Count, and Category. Blogger's name is not always extracted from XML as few sites allow bloggers to remain anonymous for posting up the blog posts even – therefore, they have been assigned as anonymous in our used dataset. The data fields specified as Comments Count does not necessarily be updated by all the blog sites' RSS. The Category data field is not always indicating the actual category as each blog site has put these categories on their own standards which is not a convention.

## 5. CONCLUSION & FUTURE WORK

The main contribution of this research is to bring into being a standard dataset in the realm of political blog mining. The dataset is specific to a single genre of blog and topic that is political blogosphere to the best of our knowledge this is very first attempt in this area. We have observed that this dataset has an immense density to be utilized for research purpose. Our applications based on this dataset is a proof of this, we have utilized this dataset into semantic search, sentiment analysis, finding influential blogger and semantic clustering of blog-post. We wish to offer this dataset for others to collaborate and research on further possibilities. In future work, we would like to pursue the questions like: whether an ordinary man's political expression can be delve into main stream politics, and hence common man participation to politics is possible or not.